%% file: sss.tex
\documentclass[runningheads]{llncs}
\usepackage{graphicx}
\usepackage{amsmath,amssymb} % define this before the line numbering.
\usepackage{color}
\usepackage{url}

\usepackage{times}
\usepackage{epsfig}
\usepackage{caption}
\usepackage{bm}
\usepackage{subfigure}
\usepackage{array}
\usepackage{tabulary}
\usepackage{multirow}
\usepackage{soul}
\usepackage{booktabs}       % professional-quality tables
\DeclareMathOperator{\prox}{\mathbf{prox}}

\begin{document}
	
% macro

\def\A{{\bf A}}
\def\a{{\bf a}}
\def\B{{\bf B}}
\def\bb{{\bf b}}
\def\C{{\bf C}}
\def\D{{\bf D}}
\def\dd{{\bf d}}
\def\E{{\bf E}}
\def\e{{\bf e}}
\def\F{{\bf F}}
\def\f{{\bf f}}
\def\G{{\bf G}}
\def\g{{\bf g}}
\def\k{{\bf k}}
\def\K{{\bf K}}
\def\H{{\bf H}}
\def\h{{\bf h}}
\def\I{{\bf I}}
\def\L{{\bf L}}
\def\M{{\bf M}}
\def\m{{\bf m}}
\def\n{{\bf n}}
\def\N{{\bf N}}
\def\BP{{\bf P}}
\def\R{{\bf R}}
\def\r{{\bf r}}
\def\BS{{\bf S}}
\def\s{{\bf s}}
\def\t{{\bf t}}
\def\T{{\bf T}}
\def\U{{\bf U}}
\def\u{{\bf u}}
\def\V{{\bf V}}
\def\v{{\bf v}}
\def\W{{\bf W}}
\def\w{{\bf w}}
\def\X{{\bf X}}
\def\x{{\bf x}}
\def\Y{{\bf Y}}
\def\Q{{\bf Q}}
\def\x{{\bf x}}
\def\y{{\bf y}}
\def\Z{{\bf Z}}
\def\z{{\bf z}}
\def\0{{\bf 0}}
\def\1{{\bf 1}}
\def\SS{{\bf S}}
\def\ME{{\mathbb E}}
\def\MA{{\mathcal A}}
\def\MF{{\mathcal F}}
\def\MG{{\mathcal G}}
\def\MI{{\mathcal I}}
\def\ML{{\mathcal L}}
\def\MN{{\mathcal N}}
\def\MO{{\mathcal O}}
\def\MT{{\mathcal T}}
\def\MX{{\mathcal X}}
\def\SW{{\mathcal {SW}}}
\def\MW{{\mathcal W}}
\def\MY{{\mathcal Y}}
\def\MR{{\mathcal R}}
\def\MF{{\mathcal F}}
\def\ML{{\mathcal L}}
\def\MG{{\mathcal G}}
\def\MH{{\mathcal H}}
\def\BR{{\mathbb R}}
\def\MS{{\mathcal S}}
\def\MC{{\mathcal C}}
\def\ph{\mbox{\boldmath$\phi$\unboldmath}}
\def\vp{\mbox{\boldmath$\varphi$\unboldmath}}
\def\pii{\mbox{\boldmath$\pi$\unboldmath}}
\def\Ph{\mbox{\boldmath$\Phi$\unboldmath}}
\def\pss{\mbox{\boldmath$\psi$\unboldmath}}
\def\Ps{\mbox{\boldmath$\Psi$\unboldmath}}
\def\muu{\mbox{\boldmath$\mu$\unboldmath}}
\def\Si{\mbox{\boldmath$\Sigma$\unboldmath}}
\def\lam{\mbox{\boldmath$\lambda$\unboldmath}}
\def\Lam{\mbox{\boldmath$\Lambda$\unboldmath}}
\def\Gam{\mbox{\boldmath$\Gamma$\unboldmath}}
\def\Oma{\mbox{\boldmath$\Omega$\unboldmath}}
\def\De{\mbox{\boldmath$\Delta$\unboldmath}}
\def\de{\mbox{\boldmath$\delta$\unboldmath}}
\def\Tha{\mbox{\boldmath$\Theta$\unboldmath}}
\def\tha{\mbox{\boldmath$\theta$\unboldmath}}
\def\etal{{\em et al.\/}\,}
\def\tr{\mathrm{tr}}
\def\exp{\mathrm{exp}}
\def\rank{\mathrm{rank}}
\def\diag{\mathrm{diag}}
\def\dg{\mathrm{dg}}
\def\argmax{\mathop{\rm argmax}}
\def\argmin{\mathop{\rm argmin}}
\def\vecd{\mathrm{vec}}
\def\bxi{\mbox{\boldmath$\xi$\unboldmath}}
\newcommand{\row}[2] {#1^{#2 \cdot}}
\newcommand{\col}[2] {#1^{\cdot #2}}
\newcommand{\norm}[1] {\|#1\|_2}
\newcommand{\normed}[1] {\frac{#1}{\|#1\|_2}}

\title{Data-Driven Sparse Structure Selection for Deep Neural Networks} 
% Replace with your title

\titlerunning{Data-Driven Sparse Structure Selection for Deep Neural Networks}
% Replace with a meaningful short version of your title
%
\author{Zehao Huang\orcidID{0000-0003-1653-208X} \and
Naiyan Wang\orcidID{0000-0002-0526-3331}}
%
%Please write out author names in full in the paper, i.e. full given and family names. 
%If any authors have names that can be parsed into FirstName LastName in multiple ways, please include the correct parsing, in a comment to the volume editors:
%\index{Lastnames, Firstnames}
%(Do not uncomment it, because you may introduce extra index items if you do that, we will use scripts for introducing index entries...)
\authorrunning{Z. Huang, N. Wang}
% Replace with shorter version of the author list. If there are more authors than fits a line, please use A. Author et al.
%

\institute{TuSimple \\
\email{\{zehaohuang18,winsty\}@gmail.com}
}
\maketitle              % typeset the header of the contribution

%%%%%%%%% ABSTRACT
\input{./sections/0-Abstract}
%%%%%%%%% INTRODUCTION
\input{./sections/1-Introduction}

%%%%%%%%% RELATED WORKS
\input{./sections/2-RelatedWorks}
%%%%%%%%% PROPOSED METHODS
\vspace{-2mm}
\input{./sections/3-ProposedMethod}
%%%%%%%%% EXPERIMENTS
\input{./sections/4-Experiments}
%%%%%%%%% CONCLUSIONS
\vspace{-3mm}
\input{./sections/6-Conclusions}
\clearpage

\bibliographystyle{splncs04}
\bibliography{modelcom&acc}
\end{document}

%% file: sections/0-Abstract.tex
\begin{abstract}
Deep convolutional neural networks have liberated its extraordinary power on various tasks. However, it is still very challenging to deploy state-of-the-art models into real-world applications due to their high computational complexity. How can we design a compact and effective network without massive experiments and expert knowledge? In this paper, we propose a simple and effective framework to learn and prune deep models in an end-to-end manner. In our framework, a new type of parameter -- scaling factor is first introduced to scale the outputs of specific structures, such as neurons, groups or residual blocks. Then we add sparsity regularizations on these factors, and solve this optimization problem by a modified stochastic Accelerated Proximal Gradient (APG) method. By forcing some of the factors to zero, we can safely remove the corresponding structures, thus prune the unimportant parts of a CNN. Comparing with other structure selection methods that may need thousands of trials or iterative fine-tuning, our method is trained fully end-to-end in one training pass without bells and whistles. We evaluate our method, Sparse Structure Selection with several state-of-the-art CNNs, and demonstrate very promising results with adaptive depth and width selection. Code is available at: \url{https://github.com/huangzehao/sparse-structure-selection}.
\keywords{sparse \and model acceleration \and deep network structure learning}
\end{abstract}

%% file: sections/1-Introduction.tex
\section{Introduction}
\label{sec:introduction}
Deep learning methods, especially convolutional neural networks (CNNs) have achieved remarkable performances in many fields, such as computer vision, natural language processing and speech recognition. However, these extraordinary performances are at the expense of high computational and storage demand. Although the power of modern GPUs has skyrocketed in the last years, these high costs are still prohibitive for CNNs to deploy in latency critical applications such as self-driving cars and augmented reality, etc.

Recently, a significant amount of works on accelerating CNNs at inference time have been proposed. Methods focus on accelerating pre-trained models include direct pruning \cite{han2015learning,li2016pruning,molchanovpruning,he2017channel,luo2017thinet}, low-rank decomposition \cite{denton2014exploiting,jaderberg2014speeding,zhang2015efficient}, and quantization \cite{rastegari2016xnor,courbariaux2016binarized,wu2016quantized}. Another stream of researches trained small and efficient networks directly, such as knowledge distillation \cite{hinton2015distilling,romero2014fitnets,Zagoruyko2017AT}, novel architecture designs \cite{iandola2016squeezenet,howard2017mobilenets} and sparse learning \cite{liu2015sparse,zhou2016less,alvarez2016learning,wen2016learning}. %Among them, sparsity based methods have attracted more attention. 
In spare learning, prior works \cite{liu2015sparse} pursued the sparsity of weights. However, non-structure sparsity only produce random connectivities and can hardly utilize current off-the-shelf hardwares such as GPUs to accelerate model inference in wall clock time. To address this problem, recently methods \cite{zhou2016less,alvarez2016learning,wen2016learning} proposed to apply group sparsity to retain a hardware friendly CNN structure.

In this paper, we take another view to jointly learn and prune a CNN. First, we introduce a new type of parameter -- scaling factors which scale the outputs of some specific structures (e.g., neurons, groups or blocks) in CNNs. These scaling factors endow more flexibility to CNN with very few parameters. Then, we add sparsity regularizations on these scaling factors to push them to zero during training. Finally, we can safely remove the structures correspond to zero scaling factors and get a pruned model. Comparing with direct pruning methods, this method is data driven and fully end-to-end. In other words, the network can select its unique configuration based on the difficulty and needs of each task. Moreover, the model selection is accomplished jointly with the normal training of CNNs. We do not require extra fine-tuning or multi-stage optimizations, and it only introduces minor cost in the training. %Comparing to group sparsity based methods, the optimization of our method is pretty simple and only introduces minimum extra computation burden in training. Besides, the learning of our method is single-shot, thus we do not need a pretrained model and we can yield lossless performance without added fine-tuning. Furthermore, our new constraint can regularize the CNNs to achieve better generalization performances. 

To summarize, our contributions are in the following three folds:

\begin{itemize}
	\item We propose a unified framework for model training and pruning in CNNs. Particularly, we formulate it as a joint sparse regularized optimization problem by introducing scaling factors and corresponding sparse regularizations on certain structures of CNNs.
	\item We utilize a modified stochastic Accelerated Proximal Gradient (APG) method to jointly optimize the weights of CNNs and scaling factors with sparsity regularizations. Compared with previous methods that utilize heuristic ways to force sparsity, our methods enjoy more stable convergence and better results without fine-tuning and multi-stage optimization.
	\item We test our proposed method on several state-of-the-art networks, PeleeNet, VGG, ResNet and ResNeXt to prune neurons, residual blocks and groups, respectively. We can adaptively adjust the depth and width accordingly. We show very promising acceleration performances on CIFAR and large scale ILSVRC 2012 image classification datasets. %Particularly, we are the first to demonstrate effective acceleration results on large scale ILSVRC dataset with state-of-the-art CNN models such as ResNet and ResNeXt.
\end{itemize}

%% file: sections/2-RelatedWorks.tex
\section{Related Works}
\label{sec:relatedworks}
\textbf{Network pruning} was pioneered in the early development of neural network. In Optimal Brain Damage \cite{lecun1989optimal} and Optimal Brain Surgeon \cite{hassibi1993second}, unimportant connections are removed based on the Hessian matrix derived from the loss function. %However, this method requires additional computation for second order derivative information. 
Recently, Han \etal \cite{han2015learning} brought back this idea by pruning the weights whose absolute value are smaller than a given threshold. This approach requires iteratively pruning and fine-tuning which is very time-consuming. To tackle this problem, Guo \etal\cite{guo2016dynamic} proposed dynamic network surgery to prune parameters during training. However, the nature of irregular sparse weights make them only yield effective compression but not faster inference in terms of wall clock time. To tackle this issue, several works pruned the neurons directly \cite{hu2016network,li2016pruning,molchanovpruning} by evaluating neuron importance on specific criteria. These methods all focus on removing the neurons whose removal affect the final prediction least. On the other hand, the diversity of neurons to be kept is also an important factor to consider \cite{mariet2015diversity}. %They used Determinantal Point Process \cite{hough2006determinantal} to select a subset of diverse neurons and subsequently merged similar neurons.
More recently, \cite{luo2017thinet} and \cite{he2017channel} formulate pruning as a optimization problem. They first select most representative neurons and further minimize the reconstitution error to recover the accuracy of pruned networks. While neuron level pruning can achieve practical acceleration with moderate accuracy loss, it is still hard to implement them in an end-to-end manner without iteratively pruning and retraining. Very recently, Liu \etal \cite{liu2017learning} used similar technique as ours to prune neurons. They sparsify the scaling parameters of batch normalization (BN) \cite{ioffe2015batch} to select channels. Ye \etal \cite{ye2018rethinking} also adopted this idea into neuron pruning. As discussed later, both of their works can be seen as a special case in our framework.

\textbf{Model structure learning} for deep learning models has attracted increasing attention recently. Several methods have been explored to learn CNN architectures without handcrafted design \cite{baker2016designing,zoph2016neural,real2017large}. One stream is to explore the design space by reinforcement learning \cite{baker2016designing,zoph2016neural} or genetic algorithms \cite{real2017large,xie2017genetic}. Another stream is to utilize sparse learning or binary optimization. 
%To automatically adapts model structure to different training data, Feng \etal \cite{feng2015learning} proposed ibpCNN to learn the number of filters in a DNN.
\cite{zhou2016less,alvarez2016learning} added group sparsity regularizations on the weights of neurons and sparsified them in the training stage.
Lately, Wen \etal \cite{wen2016learning} proposed a more general approach, which applied group sparsity on multiple structures of networks, including filter shapes, channels and layers in skip connections. Srinivas \etal \cite{srinivas2015learning} proposed a new trainable activation function  tri-state ReLU into deep networks. They pruned neurons by forcing the parameters of tri-state ReLU into binary.
%However, all these methods confront two issues: First, additional group sparse regularization may adversely deteriorate the performance. Second, the optimization of these methods are heuristic or immature. They cannot benefit from the latest developments in stochastic optimization. %Expect for reducing computation cost, their Structured Sparsity Learning (SSL) also regularized the network structure to improve performances.

\textbf{CNNs with skip connections} have been the main stream for modern network design since it can mitigate the gradient vanishing/exploding issue in ultra deep networks by the help of skip connections~\cite{srivastava2015highway,he2016deep}. Among these work, ResNet and its variants \cite{he2016identity,xie2016aggregated} have attracted more attention because of their simple design principle and state-of-the-art performances. 
%With the help of shortcuts across layers, the message propagation will not be cut off when skipping the computation of residual blocks. Several work have been explored based on it. 
%Huang \etal \cite{huang2016deep} proposed stochastic depth ResNets to enhance the training of very deep ResNets. They dropped residual blocks randomly during training and bypassed informations through skip connections. 
Recently, Veit \etal \cite{veit2016residual} interpreted ResNet as an exponential ensemble of many shallow networks. %Their explanation is also evidenced by subsequent work in network designs~\cite{wu2016wider,wang2016deeply}. 
They find there is minor impact on the performance when removing single residual block. However, deleting more and more residual blocks will impair the accuracy significantly. Therefore, accelerating this state-of-the-art network architecture is still a challenging problem. In this paper, we propose a data-driven method to learn the architecture of such kind of network. Through scaling and pruning residual blocks during training, our method can produce a more compact ResNet with faster inference speed and even better performance.  

%Recently, several works on ResNets have appeared. Shen \etal \cite{shen2016weighted} proposed Weighted Residual Networks by introducing a weight scalar into each residual block. Huang \etal \cite{huang2016deep} proposed stochastic depth ResNets to improving the training of ResNets by randomly drop layers. Both of them can train extreme deep networks successfully. Different from increasing depth, Zagoruyko \etal \cite{zagoruyko2016wide} and Wu \etal \cite{wu2016wider} focus on depth. They showed that shallower but wider ResNets outperforms much deeper models in several classification dataset. More recently, Xie \etal \cite{xie2016aggregated} exposed a new dimension, ``cardinality``, and argued that increasing it was more effective than going deeper or wider. \footnote{resnet is ensemble, inception}

%% file: sections/3-ProposedMethod.tex
\section{Proposed Method}
\label{sec:proposedmethod}
\textbf{Notations} Consider the weights of a convolutional layer $l$ in a $L$ layers CNN as a 4-dimensional tensor $\W^{l} \in \BR^{N_l\times M_l\times H_l\times W_l}$, where $N_l$ is the number of output channels, $M_l$ represents the number of input channels, $H_l$ and $W_l$ are the height and width of a 2-dimensional kernel. Then we can use $\W^{l}_{k}$ to denote the weights of $k$-th neuron in layer $l$. The scaling factors are represented as a 1-dimensional vector $\lam\in\BR^s$, where $S$ is the number of structures we consider to prune. $\lambda^i$ refers to the $i$-th value of $\lam$. Denote soft-threshold operator as $\MS_\alpha(\z)_i=\text{sign}(z_i)(|z_i|-\alpha)_{+}$.
\begin{figure*}[t]
	\centering\includegraphics[height=1.8in]{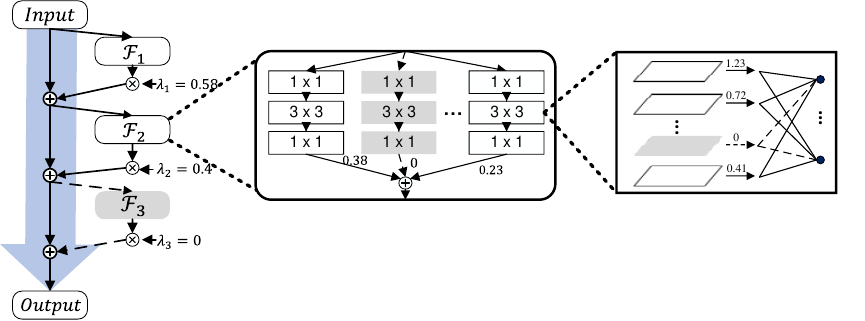}
	\caption{The network architecture of our method. $\MF$ represents a residual function. Gray block, group and neuron mean they are inactive and can be pruned since their corresponding scaling factors are $0$.}
	\label{fig:sss}
\end{figure*}  
\subsection{Sparse Structure Selection}
Given a training set consisting of $N$ sample-label pairs $\{\x_i,\y_i\}_{1\leq i\leq N}$, then a $L$ layers CNN can be represented as a function $\MC(\x_i,\W)$, where $\W=\{\W^{l}\}_{1\leq l\leq L}$ represents the collection of all weights in the CNN. $\W$ is learned through solving an optimization problem of the form:
\begin{equation}\label{equ:cnn}
\min_\W{\frac{1}{N}\sum_{i=1}^{N}{\ML(\y_i,\MC(\x_i,\W))}+\MR(\W)},
\end{equation}
where $\ML(\y_i,\MC(\x_i,\W))$ is the loss on the sample $\x_i$,  $\MR(\cdot)$ is a non-structured regularization applying on every weight, \textit{e.g.} $l_2$-norm as weight decay.

Prior sparse based model structure learning work \cite{zhou2016less,alvarez2016learning} tried to learn the number of neurons in a CNN. To achieve this goal, they added group sparsity regularization $\MR_g(\cdot)$ on $\W^{l}_k$ into Eqn.\ref{equ:cnn}, and enforced entire $\W^{l}_k$ to zero during training. Another concurrent work by Wen \etal \cite{wen2016learning} adopted similar method but on multiple different structures. %For example, in CNNs with skip connections, group sparsity can be added on $\W^{(l)}$ to cut off the connections between . 
These ideas are straightforward but the implementations are nontrivial. First, the optimization is difficult since there are several constraints on weights simultaneously, including weight decay and group sparsity. Improper optimization technique may result in slow convergence and inferior results. Consequently, there is no successful attempt to directly apply these methods on large scale applications with complicated modern network architectures.

%Additionally, the computation burden of weights' gradients respect to the sparsity regularization is overload when solving the optimization problem by Proximal Gradient Descent (PGD). In practice, both Zhou \cite{zhou2016less} and Alvarez \cite{alvarez2016learning} applied the optimization operator of sparse term at the end of each epoch instead of each iteration to avoid this problem.\footnote{To be revised.}

In this paper, we address structure learning problem in a more simple and effective way. Different from directly pushing weights in the same group to zero, we try to enforce the output of the group to zero. To achieve this goal, we introduce a new type of parameter -- scaling factor $\lam$ to scale the outputs of some specific structures (neurons, groups or blocks), and add sparsity constraint on $\lam$ during training. Our goal is to obtain a sparse $\lam$. Namely, if $\lambda^i=0$, then we can safely remove the corresponding structure since its outputs have no contribution to subsequent computation. Fig. \ref{fig:sss} illustrates our framework.

Formally, the objective function of our proposed method can be formulated as:
\begin{equation}\label{equ:sss}
\min_{\W,\lam}{\frac{1}{N}\sum_{i=1}^{N}{\ML(\y_i,\MC(\x_i,\W,\lam))}+\MR(\W)+\MR_s(\lam)},
\end{equation}
where $\MR_s(\cdot)$ is a sparsity regularization for $\lam$ with weight $\gamma$. In this work, we consider its most commonly used convex relaxation $l_1$-norm, which defined as $\gamma\|\lam\|_1$.

For $\W$, we can update it by Stochastic Gradient Descent (SGD) with momentum or its variants. For $\lam$, we adopt Accelerated Proximal Gradient (APG) \cite{parikh2014proximal} method to solve it. 
%alternatively optimizing over $\W$ and $\lam$ while keeping the other fixed. When $\lam$ is fixed, the problem of Equ. \ref{equ:sss} reduces to a general CNN optimizing problem as Equ. \ref{equ:cnn} that can be solved efficiently using SGD. When $\W$ is fixed, we adopt a proximal gradient descent (PGD) approach \cite{parikh2014proximal} to solve $\lam$.
For better illustration, we shorten $\frac{1}{N}\sum_{i=1}^{N}{\ML(\y_i,\MC(\x_i,\lam))}$ as $\MG(\lam)$, and reformulate the optimization of $\lam$ as:
\begin{equation}
\min_{\lam}{\MG(\lam)+\MR_s(\lam)}.
\end{equation}
Then we can update $\lam$ by APG:
\begin{align}
\label{equ:d}\dd_{(t)}&=\lam_{(t-1)}+\frac{t-2}{t+1}(\lam_{(t-1)}-\lam_{(t-2)})\\
\label{equ:z}\z_{(t)}&=\dd_{(t)}-\eta_{(t)}\nabla \MG(\dd_{(t)})\\
\lam_{(t)}&=\prox_{\eta_{(t)} \MR_s}(\z_{(t)}),
\end{align}
where $\eta_{(t)}$ is gradient step size at iteration $t$ and $\prox_{\eta \MR_s}(\cdot)=\MS_{\eta \gamma}(\cdot)$ since $\MR_s(\lam)=\gamma\|\lam\|_1$.
However, this formulation is not friendly for deep learning since additional to the pass for updating $\W$, we need to obtain $\nabla \MG(\dd_{(t)})$ by extra forward-backward computation, which is computational expensive for deep neural networks. 
Thus, following the derivation in \cite{sutskever2013importance}, we reformulate APG as a momentum based method:
\begin{align}
\z_{(t)}&=\lam_{(t-1)}+\mu_{(t-1)}\v_{(t-1)}\notag \\
		&~~~~-\eta_{(t)}\nabla \MG(\lam_{(t-1)}+\mu_{(t-1)}\v_{(t-1)})\\
\v_{(t)}&=\MS_{\eta_{(t)} \gamma}(\z_{(t)})-\lam_{(t-1)}\\
\lam_{(t)}&=\lam_{(t-1)}+\v_{(t)},
%\v_t&=\MS_{\eta_t \gamma}(\z_t)-\lam_{t-1}
%\v_{t_i}&=\left\{
%\begin{aligned}
%&\mu_{t-1}\v_{(t-1)_i}-\eta_{t}\nabla \MG(\lam_{t-1}+\mu_{t-1}\v_{t-1})_i - \eta_t \gamma, \quad&\text{if}&\quad \z_{t_i}>\eta_t \gamma \\
%&- \lam_{{(t-1)}_i}, \quad&\text{if}&\quad |\z_{t_i}|\leq\eta_t \gamma \\
%&\mu_{t-1}\v_{(t-1)_i}-\eta_{t}\nabla \MG(\lam_{t-1}+\mu_{t-1}\v_{t-1})_i + \eta_t \gamma, \quad&\text{if}&\quad \z_{t_i}<-\eta_t \gamma\\
%\end{aligned}
%\right.\\
\end{align}
where we define $\v_{(t-1)} = \lam_{(t-1)} - \lam_{(t-2)}$ and $\mu_{(t-1)} = \frac{t-2}{t+1}$.
This formulation is similar as the modified Nesterov Accelerated Gradient (NAG) in \cite{sutskever2013importance} except the update of $\v_t$. Furthermore, we simplified the update of $\lam$ by replacing $\lam_{(t-1)}$ as $\lam'_{(t-1)}=\lam_{(t-1)}+\mu_{(t-1)}\v_{(t-1)}$ following the modification of NAG in \cite{bengio2013advances} which has been widely used in practical deep learning frameworks \cite{chen2015mxnet}. Our new parameters $\lam'_{t}$ updates become:
\begin{align}
	\z_{(t)}&=\lam'_{(t-1)}-\eta_{(t)}\nabla \MG(\lam'_{(t-1)})\\
	\v_{(t)}&=\MS_{\eta_{(t)} \gamma}(\z_{(t)})-\lam'_{(t-1)}+\mu_{(t-1)}\v_{(t-1)}\\
	\lam'_{(t)}&=\MS_{\eta_{(t)} \gamma}(\z_{(t)})+\mu_{(t)}\v_{(t)}
\end{align}
In practice, we follow a stochastic approach with mini-batches and set momentum $\mu$ fixed to a constant value. Both $\W$ and $\lam$ are updated in each iteration. 

The implementation of APG is very simple and effective after our modification. In the following, we show it can be implemented by only ten lines of code in MXNet \cite{chen2015mxnet}.
\\
\\
\\
\noindent
{\it MXNet implementation of APG}
\begin{verbatim}
import mxnet as mx
def apg_updater(weight, lr, grad, mom, gamma):
    z = weight - lr * grad
    z = soft_thresholding(z, lr * gamma)
    mom[:] = z - weight + 0.9 * mom
    weight[:] = z + 0.9 * mom
def soft_thresholding(x, gamma):
    y = mx.nd.maximum(0, mx.nd.abs(x) - gamma)
    return mx.nd.sign(x) * y
\end{verbatim}
In our framework, we add scaling factors to three different CNN micro-structures, including neurons, groups and blocks to yield flexible structure selection. We will introduce these three cases in the following. Note that for networks with BN, we add scaling factors after BN to prevent the influence of bias parameters.
\subsection{Neuron Selection}
We introduce scaling factors for the output of channels to prune neurons. After training, removing the filters with zero scaling factor will result in a more compact network. A recent work proposed by Liu \etal \cite{liu2017learning} adopted similar idea for network slimming. They absorbed the  scaling parameters into the parameters of batch normalization, and solve the optimization by subgradient descent. During training, scaling parameters whose absolute value are lower than a threshold value are set to 0. Comparing with \cite{liu2017learning}, our method is more general and effective. Firstly, introducing scaling factor is more universal than reusing BN parameters. On one hand, some networks have no batch normalization layers, such as AlexNet \cite{krizhevsky2012imagenet} and VGG \cite{simonyan2014very}; On the other hand, when we fine-tune pre-trained models on object detection or semantic segmentation tasks, the parameters of batch normalization are usually fixed due to small batch size. Secondly, the optimization of \cite{liu2017learning} is heuristic and need iterative pruning and retraining. In contrast, our optimization is more stable in an end-to-end manner. Above all, \cite{liu2017learning} can be seen as a special case of our method. Similarly, \cite{ye2018rethinking} is also a special case of our method. The difference between Ye \etal \cite{ye2018rethinking} and Liu \etal \cite{liu2017learning} is Ye \etal adopted ISTA \cite{beck2009fast} to optimize scaling factors. We will compare these different optimization methods in our experiments.
 
\subsection{Block Selection}
The structure of skip connection CNNs allows us to skip the computation of specific layers without cutting off the information flow in the network. Through stacking residual blocks, ResNet \cite{he2016deep,he2016identity} can easily exploit the advantage of very deep networks. Formally, residual block with identity mapping can be formulated by the following formula:
\begin{equation}\label{equ:resnet}
\r^{i+1}=\r^{i}+\MF^{i}(\r^i, \W^i),
\end{equation}
where $\r^{i}$ and $\r^{i+1}$ are input and output of the $i$-th block, $\MF^i$ is a residual function and $\W^i$ are parameters of the block. 

To prune blocks, we add scaling factor after each residual block. Then in our framework, the formulation of Eqn.\ref{equ:resnet} is as follows:
\begin{equation}
\r^{i+1}=\r^{i}+\lambda^i\MF^{i}(\r^i, \W^i).
\end{equation}
As shown in Fig \ref{fig:sss}, after optimization, we can get a sparse $\lam$. The residual block with scaling factor 0 will be pruned entirely, and we can learn a much shallower ResNet.
A prior work that also adds scaling factors for residual in ResNet is Weighted Residual Networks \cite{shen2016weighted}. Though sharing a lot of similarities, the motivations behind these two works are different. Their work focuses on how to train ultra deep ResNet to get better results with the help of scaling factors. Particularly, they increase depth from 100+ to 1000+. While our method aims to decrease the depth of ResNet, we use the scaling factors and sparse regularizations to sparsify the output of residual blocks.%can achieve better performance than normal ResNets both in shallow and deep. The sparsity of $\lam$ yield better regularization than $l_2$ regularizer adopted in \cite{shen2016weighted}.

\subsection{Group Selection}
Recently, Xie \etal introduced a new dimension -- cardinality into ResNets and proposed ResNeXt \cite{xie2016aggregated}. Formally, they presented aggregated transformations as:
\begin{equation}
\MA(\x)=\sum_{i=1}^{C}\MT^i(\x,\W^i),
\end{equation}
where $\MT^i(\x)$ represents a transformation with parameters $\W^i$, $C$ is the cardinality of the set of $\MT^i(\x)$ to be aggregated. In practice, they use grouped convolution to ease the implementation of aggregated transformations. So in our framework, we refer $C$ as the number of group, and formulate a weighted $\MA(\x)$ as:
\begin{equation}
\label{equ:lambdagroup}
\MA(\x)=\sum_{i=1}^{C}\lambda^i\MT^i(\x, \W^i)
\end{equation}
After training, several basic cardinalities are chosen by a sparse $\lam$ to form the final transformations. Then, the inactive groups with zero scaling factors can be safely removed as shown in Fig \ref{fig:sss}. Note that neuron pruning can also seen as a special case of group pruning when each group contains only one neuron. Furthermore, we can combine block pruning and group pruning to learn more flexible network structures. 

%% file: sections/4-Experiments.tex
\section{Experiments}
\label{sec:experiments}
\begin{figure*}[t]
	\centering
	\subfigure[\scriptsize VGG CIFAR10]{\includegraphics[width=0.24\linewidth]{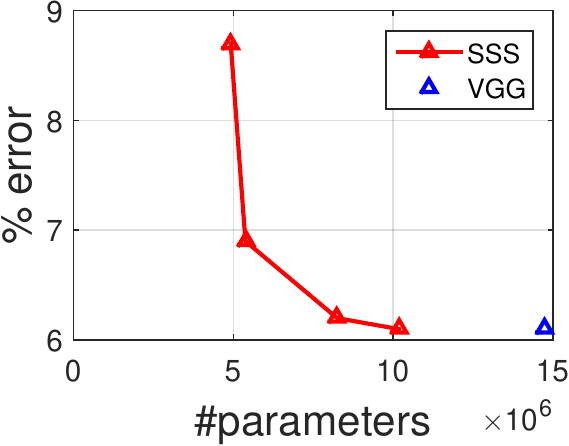}}
	\subfigure[\scriptsize VGG CIFAR10]{\includegraphics[width=0.24\linewidth]{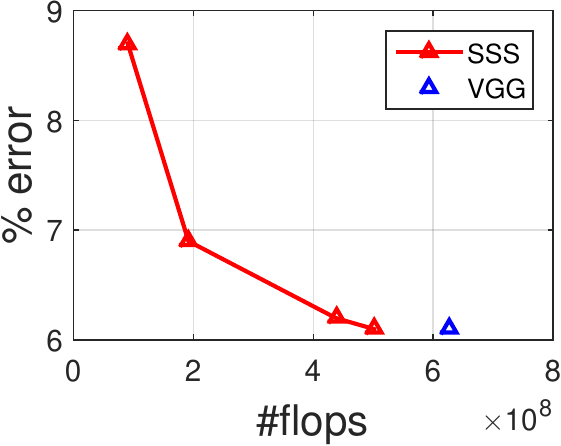}}
	\subfigure[\scriptsize VGG CIFAR100]{\includegraphics[width=0.24\linewidth]{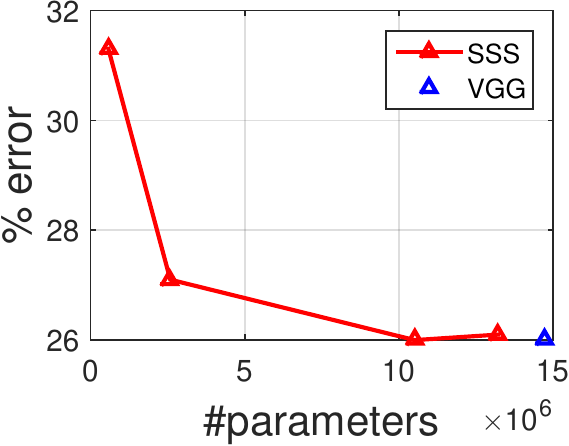}}
	\subfigure[\scriptsize VGG CIFAR100]{\includegraphics[width=0.24\linewidth]{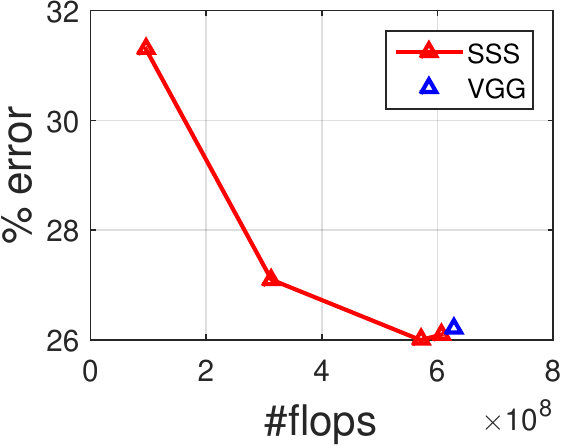}}
	\caption{Error vs. number of parameters and FLOPs after SSS training for VGG on CIFAR-10 and CIFAR-100 datasets.}\label{fig:VGGCIFAR}
\end{figure*}
\begin{figure*}[t]
	\centering
	\subfigure[\scriptsize ResNet20 CIFAR10]{\includegraphics[width=0.24\linewidth]{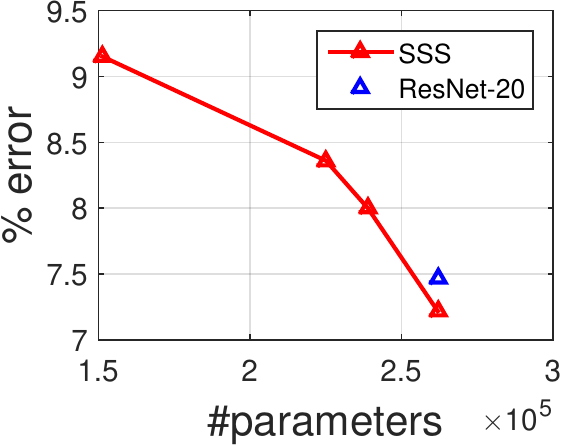}}
	\subfigure[\scriptsize ResNet20 CIFAR10]{\includegraphics[width=0.24\linewidth]{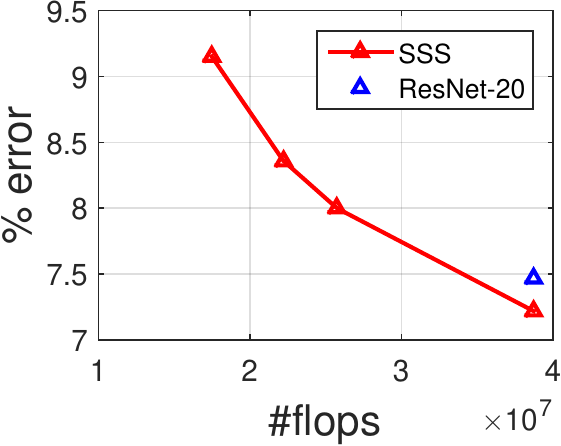}}
	\subfigure[\scriptsize ResNet20 CIFAR100]{\includegraphics[width=0.24\linewidth]{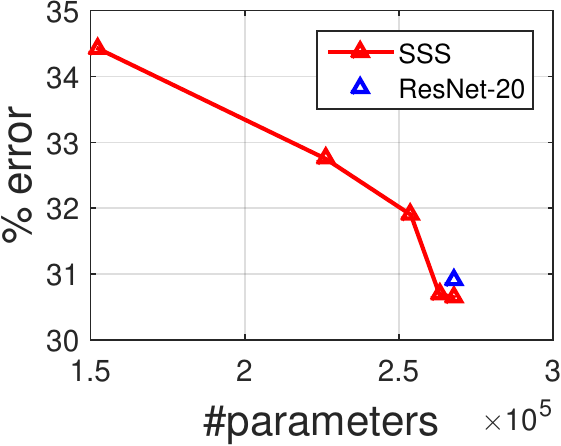}}
	\subfigure[\scriptsize ResNet20 CIFAR100]{\includegraphics[width=0.24\linewidth]{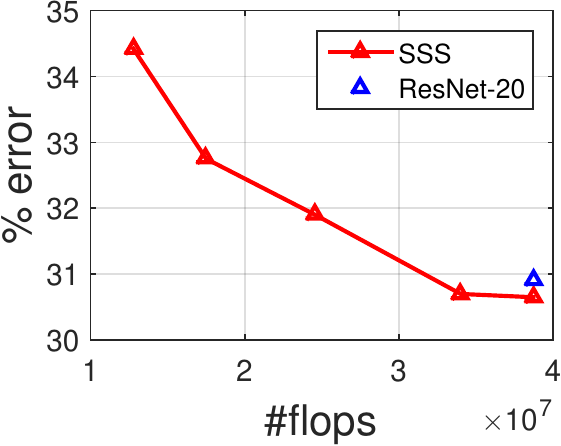}}
	\subfigure[\scriptsize ResNet164 CIFAR10]{\includegraphics[width=0.24\linewidth]{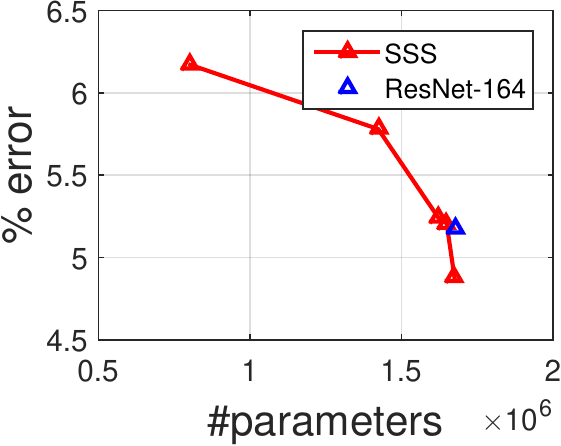}}
	\subfigure[\scriptsize ResNet164 CIFAR10]{\includegraphics[width=0.24\linewidth]{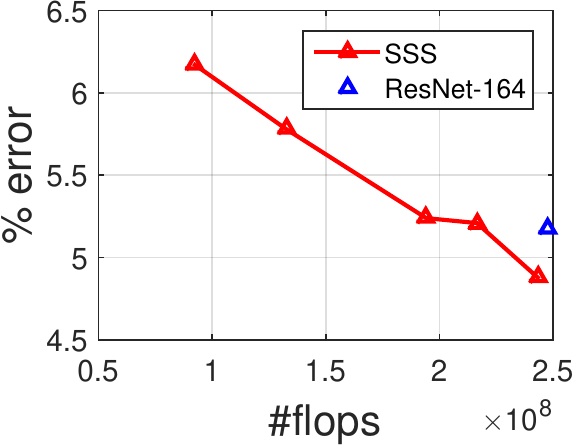}}
	\subfigure[\scriptsize ResNet164 CIFAR100]{\includegraphics[width=0.24\linewidth]{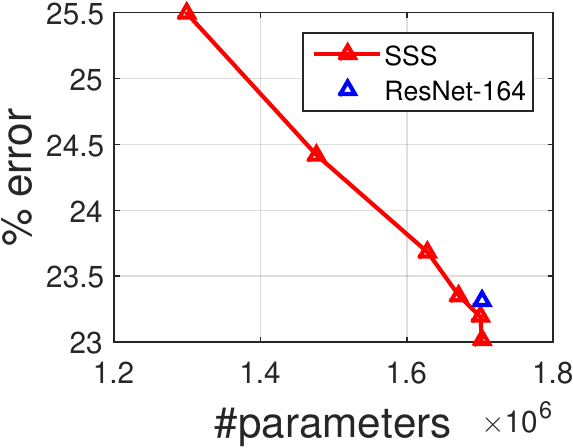}}
	\subfigure[\scriptsize ResNet164 CIFAR100]{\includegraphics[width=0.24\linewidth]{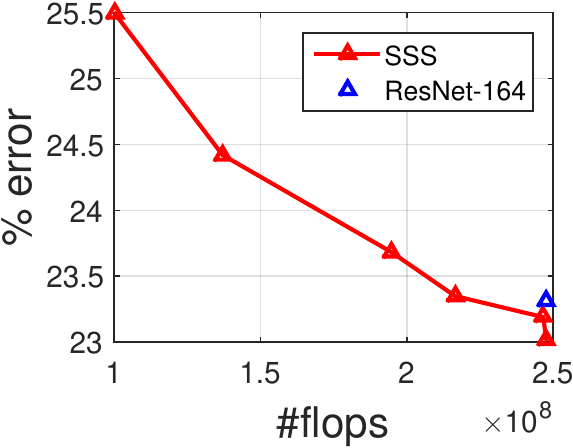}}
	\caption{Error vs. number of parameters and FLOPs after SSS training for ResNet-20 and ResNet-164 on CIFAR-10 and CIFAR-100 datasets.}\label{fig:ResNetCIFAR}
\end{figure*}
\begin{figure*}[t]
	\centering
	\subfigure[{\scriptsize ResNeXt20 CIFAR10}]{\includegraphics[width=0.24\linewidth]{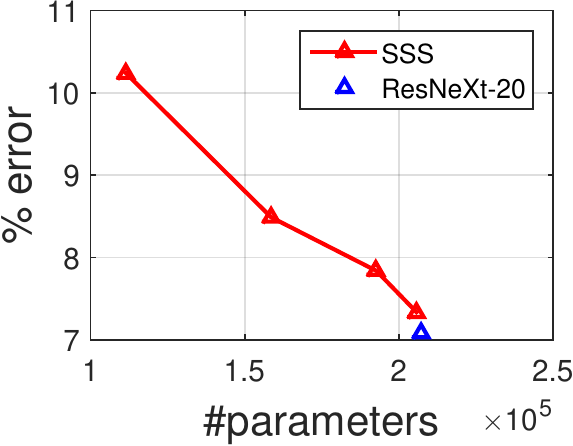}}
	\subfigure[{\scriptsize ResNeXt20 CIFAR10}]{\includegraphics[width=0.24\linewidth]{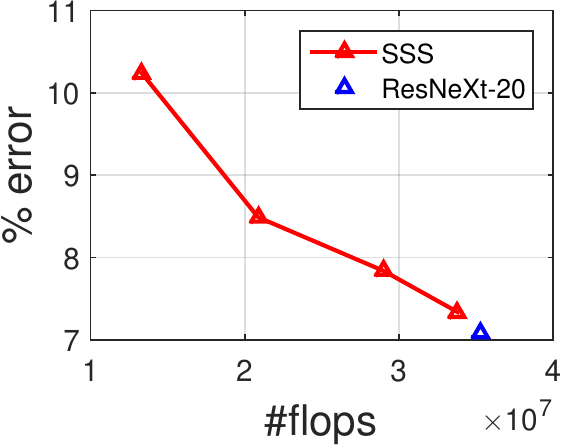}}
	\subfigure[{\scriptsize ResNeXt20 CIFAR100}]{\includegraphics[width=0.24\linewidth]{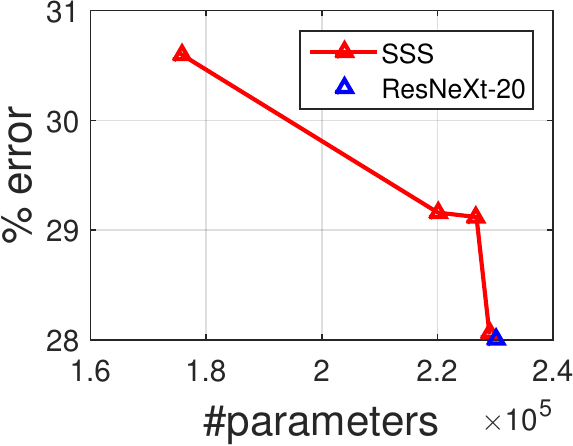}}
	\subfigure[{\scriptsize ResNeXt20 CIFAR100}]{\includegraphics[width=0.24\linewidth]{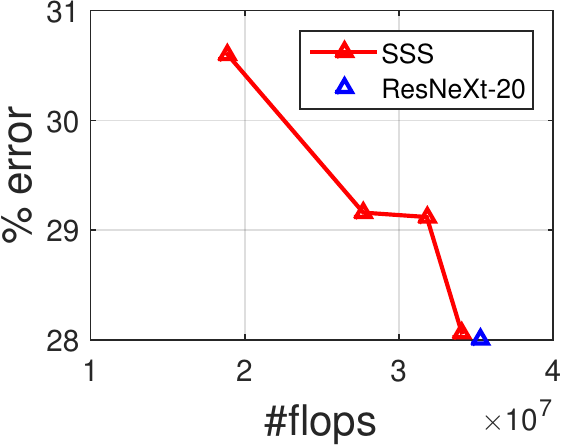}}
	\subfigure[{\scriptsize ResNeXt164 CIFAR10}]{\includegraphics[width=0.24\linewidth]{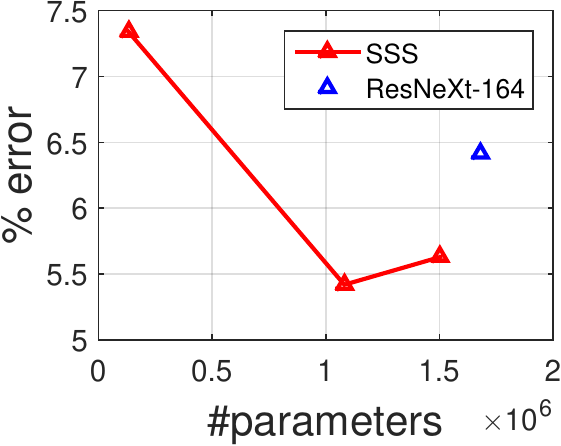}}
	\subfigure[{\scriptsize ResNeXt164 CIFAR10}]{\includegraphics[width=0.24\linewidth]{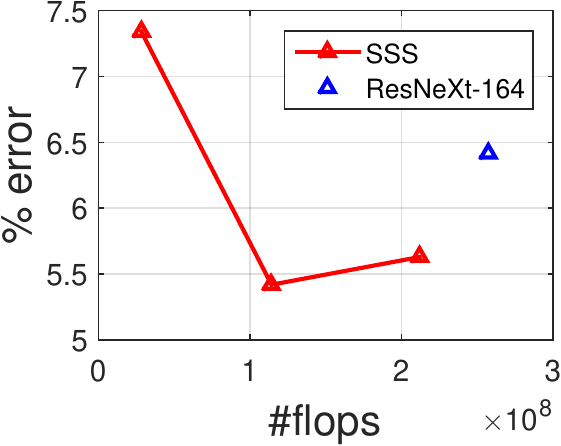}}
	\subfigure[{\scriptsize ResNeXt164 CIFAR100}]{\includegraphics[width=0.24\linewidth]{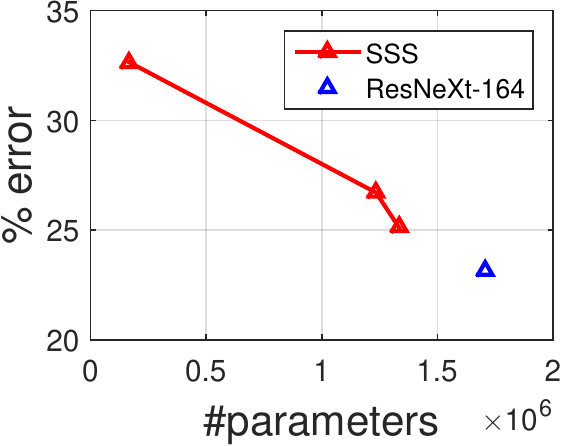}}
	\subfigure[{\scriptsize ResNeXt164 CIFAR100}]{\includegraphics[width=0.24\linewidth]{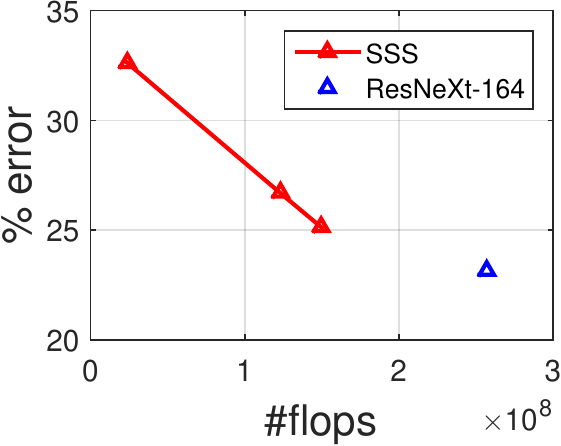}}
	\caption{Error vs. number of parameters and FLOPs with SSS training for ResNeXt-20 and ResNeXt-164 on CIFAR-10 and CIFAR-100 datasets.}\label{fig:ResNeXtCIFAR}
\end{figure*}
\begin{figure*}[h]%[b]{0.75\linewidth}
	\centering
	\subfigure[Parameters]{\includegraphics[width=0.44\linewidth]{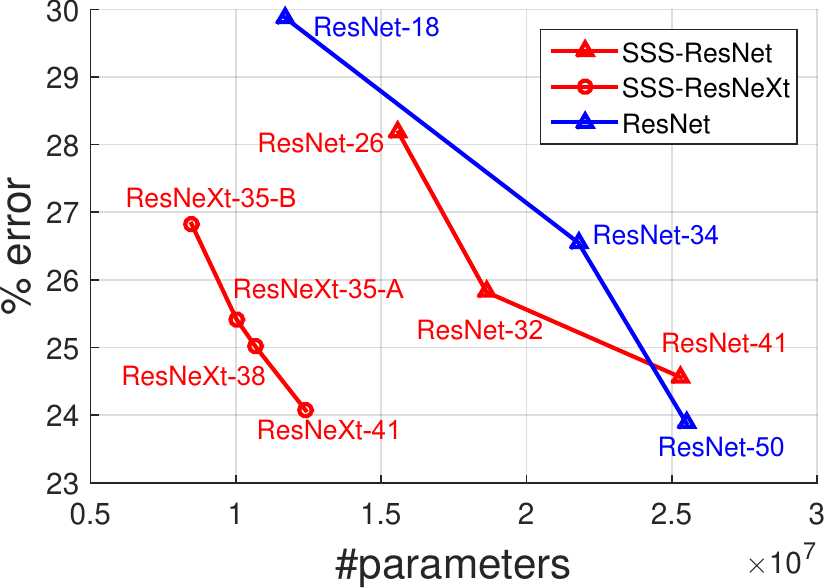}}
	\subfigure[FLOPs]{\includegraphics[width=0.44\linewidth]{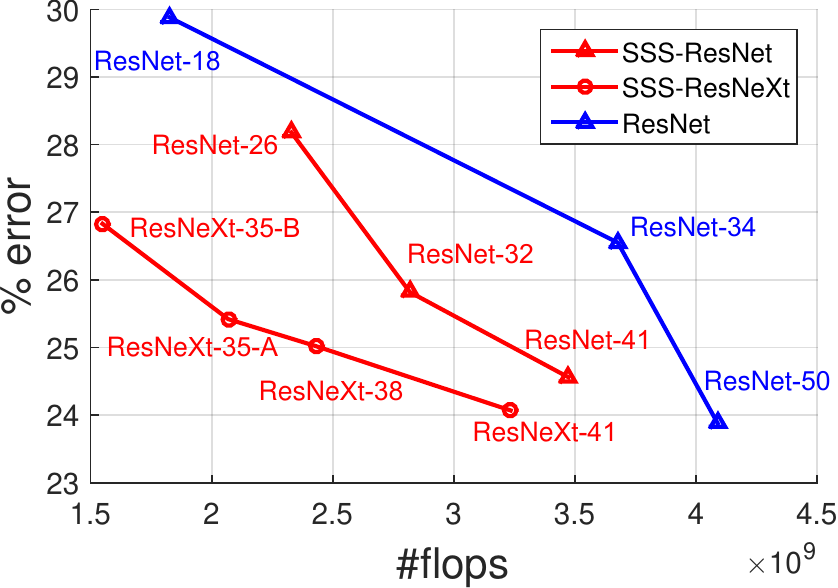}}
	\captionof{figure}{Top-1 error vs. number of parameters and FLOPs for our SSS models and original ResNets on ImageNet validation set.}\label{fig:ImageNet}
\end{figure*}
In this section, we evaluate the effectiveness of our method on three standard datasets, including CIFAR-10, CIFAR-100 \cite{krizhevsky2009learning} and ImageNet LSVRC 2012 \cite{russakovsky2015imagenet}. For neuron pruning, we adopt VGG16 \cite{simonyan2014very}, a classical plain network to validate our method. As for blocks and groups, we use two state-of-the-art networks, ResNet \cite{he2016identity} and ResNeXt \cite{xie2016aggregated} respectively. To prove the practicability of our method, we further experiment in a very lightweight network, PeleeNet \cite{wang2018pelee}. 

For optimization, we adopt NAG \cite{sutskever2013importance,bengio2013advances} and our modified APG to update weights $\W$ and scaling factors $\lam$, respectively. We set weight decay of $\W$ to 0.0001 and fix momentum to 0.9 for both $\W$ and $\lam$. The weights are initialized as in \cite{he2016deep} and all scaling factors are initialized to be 1. All the experiments are conducted in MXNet \cite{chen2015mxnet}. %The speedups are measured on a single Titan X Pascal. 
\subsection{CIFAR}
We start with CIFAR dataset to evaluate our method. CIFAR-10 dataset consists of 50K training and 10K testing RGB images with 10 classes. CIFAR-100 is similar to CIFAR-10, except it has 100 classes. 
As suggested in \cite{he2016deep}, the input image is $32\times32$ randomly cropped from a zero-padded $40\times40$ image or its flipping. The models in our experiments are trained with a mini-batch size of 64 on a single GPU. We start from a learning rate of 0.1 and train the models for 240 epochs. The learning rate is divided by 10 at the 120-th,160-th and 200-th epoch.

\textbf{VGG:}~The baseline network is a modified VGG16 with BN \cite{ioffe2015batch}\footnote{Without BN, the performance of this network is very worse in CIFAR-100 dataset.}. We remove fc6 and fc7 and only use one fully-connected layer for classification. We add scale factors after every batch normalization layers. Fig. \ref{fig:VGGCIFAR} shows the results of our method. Both parameters and floating-point operations per second (FLOPs)\footnote{Multiply-adds.} are reported. Our method can save about 30\% parameters and 30\% - 50\% computational cost with minor lost of performance.

\textbf{ResNet:}~To learn the number of residual blocks, we use ResNet-20 and ResNet-164 \cite{he2016identity} as our baseline networks. ResNet-20 consists of 9 residual blocks. Each block has 2 convolutional layers, while ResNet-164 has 54 blocks with bottleneck structure in each block. Fig. \ref{fig:ResNetCIFAR} summarizes our results. It is easy to see that our SSS achieves better performance than the baseline model with similar parameters and FLOPs. For ResNet-164, our SSS yields 2.5x speedup with about $2\%$ performance loss both in CIFAR-10 and CIFAR-100. %The architectures of three pruned ResNets for CIFAR-100 are shown in Table \ref{table:ResNet164cifar100}. 
After optimization, we found that the blocks in early stages are pruned first. This discovery coincides with the common design that the network should spend more budget in its later stage, since more and more diverse and complicated pattern may emerge as the receptive field increases.%And no downsampling residual blocks are pruned by our data driven method. This finding is consistent with the experiments in \cite{veit2016residual}. %Thus, our method can select important blocks and prune others automatically based on the training data.

\begin{table*}[htb]
	\centering	
	\caption{Network architectures of ResNet-50 and our pruned ResNets for ImageNet. $\surd$ represents that the corresponding block is kept while $\times$ denotes that the block is pruned}
	%\vspace{8pt}
	\begin{tabular}{c|c|c|c|c|c}
		\toprule
		stage                  & output                        & ResNet-50                                                                                            & ResNet-26             & \multicolumn{1}{l|}{ResNet-32} & ResNet-41 \\ \midrule
		conv1                  & 112$\times$112                & \multicolumn{4}{c}{7$\times$7,~64,~stride 2}                                                                                                                                   \\ \midrule
		\multirow{2}{*}{conv2} & \multirow{2}{*}{56$\times$56} & \multicolumn{4}{c}{3$\times$3 max pool, stride 2}                                                                                                                                        \\ \cline{3-6} 
		&                               & \multicolumn{1}{l|}{~$\begin{bmatrix}\text{1}\times\text{1},\text{64}~~\\\text{3}\times\text{3},\text{64}~~\\\text{1}\times\text{1},\text{256}\end{bmatrix}\times\text{3}$} & $\times\times\times$ & $\times\times\surd$          & $\times\times\times$                          \\ \midrule
		conv3                  & 28$\times$28                  & $\begin{bmatrix}\text{1}\times\text{1},\text{128}~~\\\text{3}\times\text{3},\text{128}~~\\\text{1}\times\text{1},\text{512}~~\end{bmatrix}\times\text{4}$                  &  $\times\surd\surd\times$                     & $\surd\surd\surd\surd$                               &  $\surd\surd\surd\surd$                         \\ \midrule
		conv4                  & 14$\times$14                  & $\begin{bmatrix}\text{1}\times\text{1},\text{256}~~\\\text{3}\times\text{3},\text{256}~~\\\text{1}\times\text{1},\text{1024}\end{bmatrix}\times\text{6}$                   &   $\times\surd\surd\surd\surd\surd$                    & $\surd\surd\surd\times\times\surd$                               & $\surd\surd\surd\surd\surd\surd$                          \\ \midrule
		conv5                  & 7$\times$7                    & $\begin{bmatrix}\text{1}\times\text{1},\text{512}~~\\\text{3}\times\text{3},\text{512}~~\\\text{1}\times\text{1},\text{2048}\end{bmatrix}\times\text{3}$                   & $\surd\times\times$                      & $\surd\times\times$                               & $\surd\surd\surd$                          \\ \midrule
		& 1$\times$1                    & \multicolumn{4}{c}{global average pool 1000-d FC, softmax}                                                                                                                               \\ \bottomrule
	\end{tabular}
	%\vspace{-8pt}
	\label{table:structure}
\end{table*}

\textbf{ResNeXt:}~We also test our method on ResNeXt \cite{xie2016aggregated}. We choose  ResNeXt-20 and ResNeXt-164 as our base networks. Both of these two networks have bottleneck structures with 32 groups in residual blocks. For ResNeXt-20, we focus on groups pruning since there are only 6 residual blocks in it. For ResNeXt-164, we add sparsity on both groups and blocks. Fig. \ref{fig:ResNeXtCIFAR} shows our experiment results. Both groups pruning and block pruning show good trade-off between parameters and performance, especially in ResNeXt-164. The combination of groups and blocks pruning is extremely effective in CIFAR-10. Our SSS saves about $60\%$ FLOPs while achieves $1\%$ higher accuracy. In ResNeXt-20, groups in first and second block are pruned first. Similarly, in ResNeXt-164, groups in shallow residual blocks are pruned mostly.

\begin{table}[!htb]
	\caption{Results on ImageNet dataset. Both top-1 and top-5 validation errors (single crop) are reported. Number of parameters and FLOPs for inference of different models are also shown. Here, M/B means million/billion ($10^6/10^9$), respectively}\label{table:imagenet}
	%\vspace{8pt}
	\centering
	\footnotesize
	\begin{tabular}{l|cccc}
		\toprule
		%\hline
		\multicolumn{1}{c|}{Model} & Top-1 & Top-5 & \#Parameters & \multicolumn{1}{c}{\#FLOPs} \\ \midrule
		VGG-16       & 27.54 & 9.16 & 138.3M& 30.97B  \\
		VGG-16       & 31.47 & 11.8 & 130.5M & 7.667B\\ \midrule
		ResNet-50    & 23.88 & 7.14  & 25.5M  & 4.089B                          \\
		ResNet-41    & 24.56 & 7.39  & 25.3M  & 3.473B                          \\ 
		%ResNet-34 (Baseline)       & 34                         & 27.56 & 9.03
		%&21.8M           & 3.677$\times10^9$                            \\
		ResNet-32    & 25.82 & 8.09 &18.6M   & 2.818B                        \\ 
		%ResNet-18 (Baseline)       & 18                         & 31.42 & 11.34
		%&11.7M            & 1.827$\times10^9$                           \\
		ResNet-26    & 28.18 & 9.21 &15.6M   & 2.329B                          \\ \midrule
		ResNeXt-50   & 22.43 & 6.32 &25.0M & 4.230B \\
		ResNeXt-41   & 24.07 & 7.00 &12.4M & 3.234B \\  
		ResNeXt-38   & 25.02 & 7.50 &10.7M & 2.431B \\ 
		ResNeXt-35-A & 25.43 & 7.83 &10.0M & 2.068B \\ 
		ResNeXt-35-B & 26.83 & 8.42 &8.50M & 1.549B \\\bottomrule
	\end{tabular}
\end{table}
\subsection{ImageNet LSVRC 2012}
%\begin{table*}[ht]
%	\begin{minipage}[b]{0.3\linewidth}
%		\tiny
%		\setlength{\tabcolsep}{4.5pt}
%		\centering
%		\begin{tabular}{l|c}
%			\toprule
%			Layer  & Width \\ \hline
%			conv11  & 18 \\
%			conv12  & 35 \\ 
%			conv21  & 87 \\
%			conv22  & 79 \\
%			conv31  & 88 \\
%			conv32  & 61 \\
%			conv33  & 79 \\
%			conv41  & 180 \\
%			conv42  & 198 \\
%			conv43  & 230 \\
%			conv51  & 512 \\
%			conv52  & 512 \\
%			conv53  & 512 \\
%			\bottomrule
%		\end{tabular}
%		\vspace{2pt}
%		\captionof{table}{Pruned VGG16 structure.}
%		\label{table:vgg-structure}
%	\end{minipage}\hfill

%\end{table*}

To further demonstrate the effectiveness of our method in large-scale CNNs, we conduct more experiments on the ImageNet LSVRC 2012 classification task with  VGG16 \cite{simonyan2014very}, ResNet-50 \cite{he2016identity} and ResNeXt-50 (32 $\times$ 4d) \cite{xie2016aggregated}. We do data augmentation based on the publicly available implementation of ``fb.resnet''~\footnote{\url{https://github.com/facebook/fb.resnet.torch}}. The mini-batch size is 128 on 4 GPUs for VGG16 and ResNet-50, and 256 on 8 GPUs for ResNeXt-50. The optimization and initialization are similar as those in CIFAR experiments. We train the models for 100 epochs. The learning rate is set to an initial value of 0.1 and then divided by 10 at the 30-th, 60-th and 90-th epoch. All the results for ImageNet dataset are summarized in Table \ref{table:imagenet}. 
%\begin{table}
%	\caption{Comparison between our ResNeXt-38 and DenseNet-121.}
%	\label{table:vgg-structure}
%	\footnotesize
%	\setlength{\tabcolsep}{4.5pt}
%	\centering
%	\begin{tabular}{|l|c|c|}
%		\hline
%		Layer & Width & Width* \\ \hline
%		conv11 & 64 & 18 \\
%		conv12 & 64 & 35 \\ 
%		conv21 & 128 & 87 \\
%		conv22 & 128 & 79 \\
%		conv31 & 256 & 88 \\
%		conv32 & 256 & 61 \\
%		conv33 & 256 & 79 \\
%		conv41 & 512 & 180 \\
%		conv42 & 512 & 198 \\
%		conv43 & 512 & 230 \\
%		conv51 & 512 & 512 \\
%		conv52 & 512 & 512 \\
%		conv53 & 512 & 512 \\
%		\hline
%	\end{tabular}
%\end{table}
%\begin{figure*}[htb]
%	\centering
%	\subfigure[Parameters]{\includegraphics[width=0.4\linewidth]{figs/imagenet_params}}
%	\subfigure[FLOPs]{\includegraphics[width=0.4\linewidth]{figs/imagenet_flops}}
%	\caption{Top-1 error vs. number of parameters and FLOPs for our SSS models and original ResNets on ImageNet validation set.}\label{fig:ImageNet}
%\end{figure*}

\textbf{VGG16:}~In our experiments of VGG16 pruning, we find the results of pruning all convolutional layers were not promising. This is because in VGG16, the computational cost in terms of FLOPs is not equally distributed in each layer. The number of FLOPs of conv5 layers is 2.77 billion in total, which is only 9\% of the whole network (30.97 billion). Thus, we consider the sparse penalty should be adjusted by computational cost of different layers. Similar idea has been adopted in \cite{molchanovpruning} and \cite{he2017channel}. In \cite{molchanovpruning}, they introduce FLOPs regularization to the pruning criteria. He \etal \cite{he2017channel} do not prune conv5 layers in their VGG16 experiments. Following \cite{he2017channel}, we set the sparse penalty of conv5 to 0 and only prune conv1 to conv4. The results can be found in Table \ref{table:imagenet}. The pruned model save about 75\% FLOPs, while the parameter saving is negligible. This is due to that fully-connected layers have a large amount of parameters (123 million in original VGG16), and we do not pruned fully-connected layers for fair comparison with other methods. %Though our result has performance degradation, in the following section, we will show it is still competitive to other state-of-the-art methods.

\textbf{ResNet-50:}~For ResNet-50, we experiment three different settings of $\gamma$ to explore the performance of our method in block pruning. For simplicity, we denote the trained models as ResNet-26, ResNet-32 and ResNet-41 depending on their depths. Their structures are shown in Table \ref{table:structure}. All the pruned models come with accuracy loss in certain extent. Comparing with original ResNet-50, ResNet-41 provides $15\%$ FLOPs reduction with $0.7\%$ top-1 accuracy loss while ResNet-32 saves $31\%$ FLOPs with about $2\%$ top-1 loss. Fig. \ref{fig:ImageNet} shows the top-1 validation errors of our SSS models and ResNets as a function of the number of parameters and FLOPs. The results reveal that our pruned models perform on par with original hand-crafted ResNets, whilst requiring less parameters and computational cost. For example, comparing with ResNet-34 \cite{he2016identity}, both our ResNet-41 and ResNet-32 yield better performances with less FLOPs.

\textbf{ResNeXt-50:}~As for ResNeXt-50, we add sparsity constraint on both residual blocks and groups which results in several pruned models. Table \ref{table:imagenet} summarizes the performance of these models. The learned ResNeXt-41 yields $24\%$ top-1 error in ILSVRC validation set. It gets similar results with the original ResNet50, but with half parameters and more than 20\% less FLOPs. In ResNeXt-41, three residual blocks in ``conv5'' stage are pruned entirely. This pruning result is somewhat contradict to the common design of CNNs, which worth to be studied in depth in the future. 
%\begin{table}
%	\caption{Comparison among several state-of-the-art pruning methods on the ResNet and VGG16 networks.}
%	\label{table:resnet-comparison}
%	\footnotesize
%	\setlength{\tabcolsep}{4.5pt}
%	\centering
%	\begin{tabular}{l|c|c|c|c}
%		\toprule
%		Model & Method & Top-1 & Top-5 & \#FLOPs  \\ \midrule
%		ResNet-34-pruned & Li \etal \cite{li2016pruning}& 27.44 & - & 3.080B  \\
%		ResNet-50-pruned-A & Li \etal \cite{li2016pruning} our impl.& 27.12 & 8.95 & 3.070B\\
%		ResNet-50-pruned-B & Li \etal \cite{li2016pruning} our impl.& 27.02 & 8.92 & 3.400B  \\ 
%		ResNet-32 & Ours & 25.82 & 8.09 & 2.818B  \\\midrule
%		ResNet-50-pruned & He \etal \cite{he2017channel} & - & 9.20 & 2.726B\\ 
%		ResNet-26 & Ours & 28.18 & 9.21 & 2.329B \\ \midrule
%		VGG16-pruned (5$\times$) & He \etal \cite{he2017channel} & 32.20 & 11.90 & 7.033B \\
%		VGG16-pruned (ThiNet-Conv) & Luo \etal \cite{luo2017thinet} & 30.20 & 10.47 & 9.580B \\
%		VGG16-pruned & Ours & 31.47 & 11.80 & 7.667B \\
%		\hline
%	\end{tabular}
%\end{table}
\subsection{Pruning lightweight network}
Adopting lightweight networks, such as MobileNet\cite{howard2017mobilenets}, ShuffleNet\cite{zhang2018shuffle} for fast inference is a more effective strategy in practice. To future prove the effectiveness of our method, we adopt neuron pruning in PeleeNet\cite{wang2018pelee}, which is a state-of-the-art efficient architecture without separable convolution. We follow the training settings and hyper-parameters used in \cite{zhang2018shuffle}. The mini-batch size is 1024 on 8 GPUs and we train 240 epoch.
Table \ref{table:pelee} shows the pruning results of PeleeNet. We adopt different settings of $\gamma$ and get three pruned networks. Comparing to baseline, Our purned PeleeNet-A save about 14\% parameters and FLOPs with only 0.4\% top-1 accuracy degradation. %Comparing to MobileNet-0.75-224 \cite{howard2017mobilenets}, our pruned PeleeNet-A achieves about 0.8\% higher top-1 accuracy with less FLOPs.
\begin{table}[t]
	\footnotesize
	\setlength{\tabcolsep}{4.5pt}
	\centering
	\caption{Results of PeleeNet on ImageNet dataset}	\label{table:pelee}
	%\vspace{8pt}
	\begin{tabular}{l|c|c|c|c}
		\toprule
		Model & Top-1 & Top-5 & \#Parameters & \#FLOPs  \\ \midrule
		PeleeNet (Our impl.)& 27.47 & 9.15 & 2.8M & 508M  \\
		PeleeNet-A & 27.85 & 9.34 & 2.4M & 436M \\
		PeleeNet-B & 30.87 & 11.38 & 1.6M & 293M \\
		PeleeNet-C & 32.81 & 12.69 & 1.4M & 236M \\
		\bottomrule
	\end{tabular}
	
\end{table}
\subsection{Comparison with other methods}
\begin{figure*}[t]
	\centering
	\subfigure[\scriptsize CIFAR10]{\includegraphics[width=0.3\linewidth]{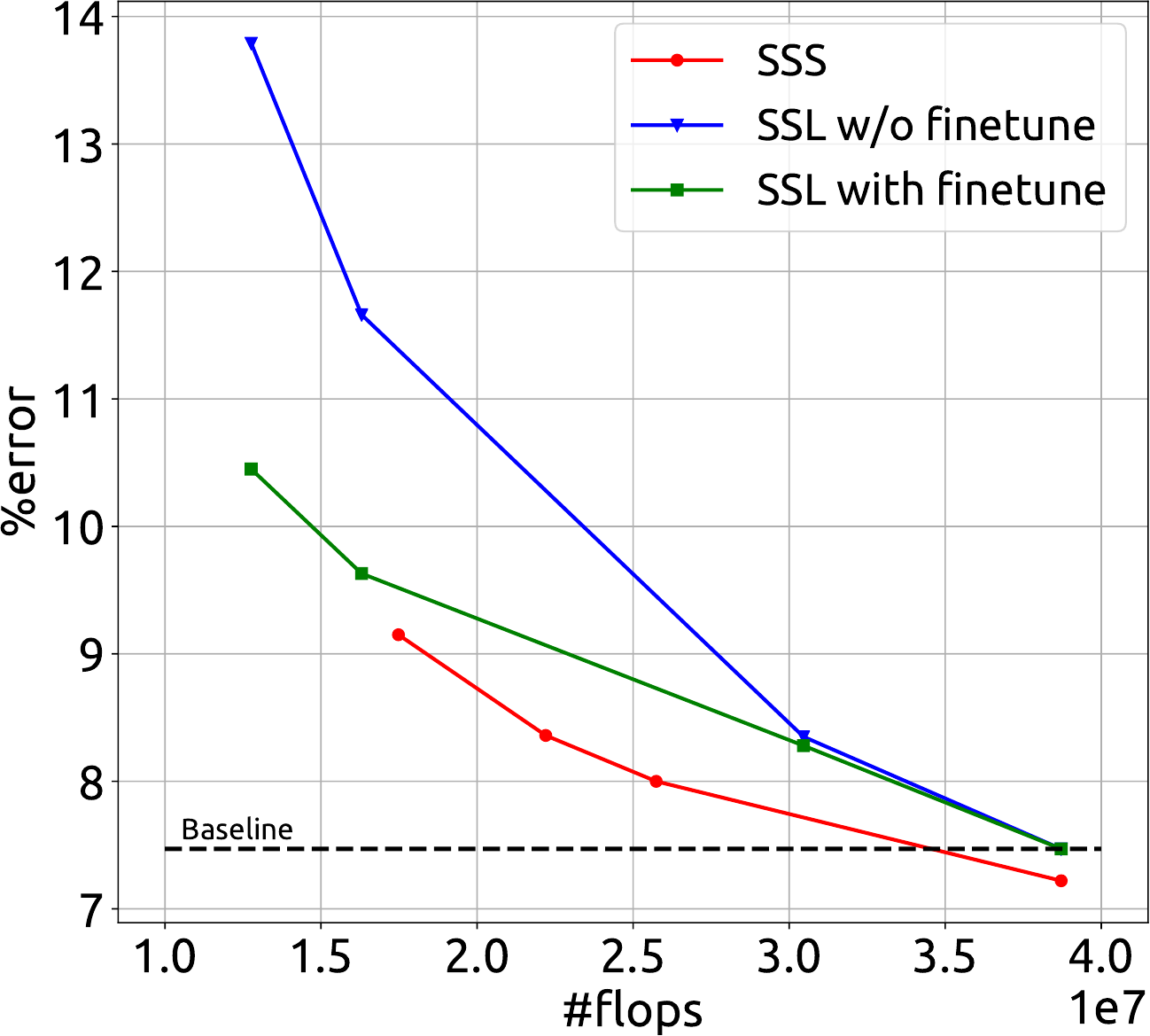}}
	\subfigure[\scriptsize CIFAR100]{\includegraphics[width=0.3\linewidth]{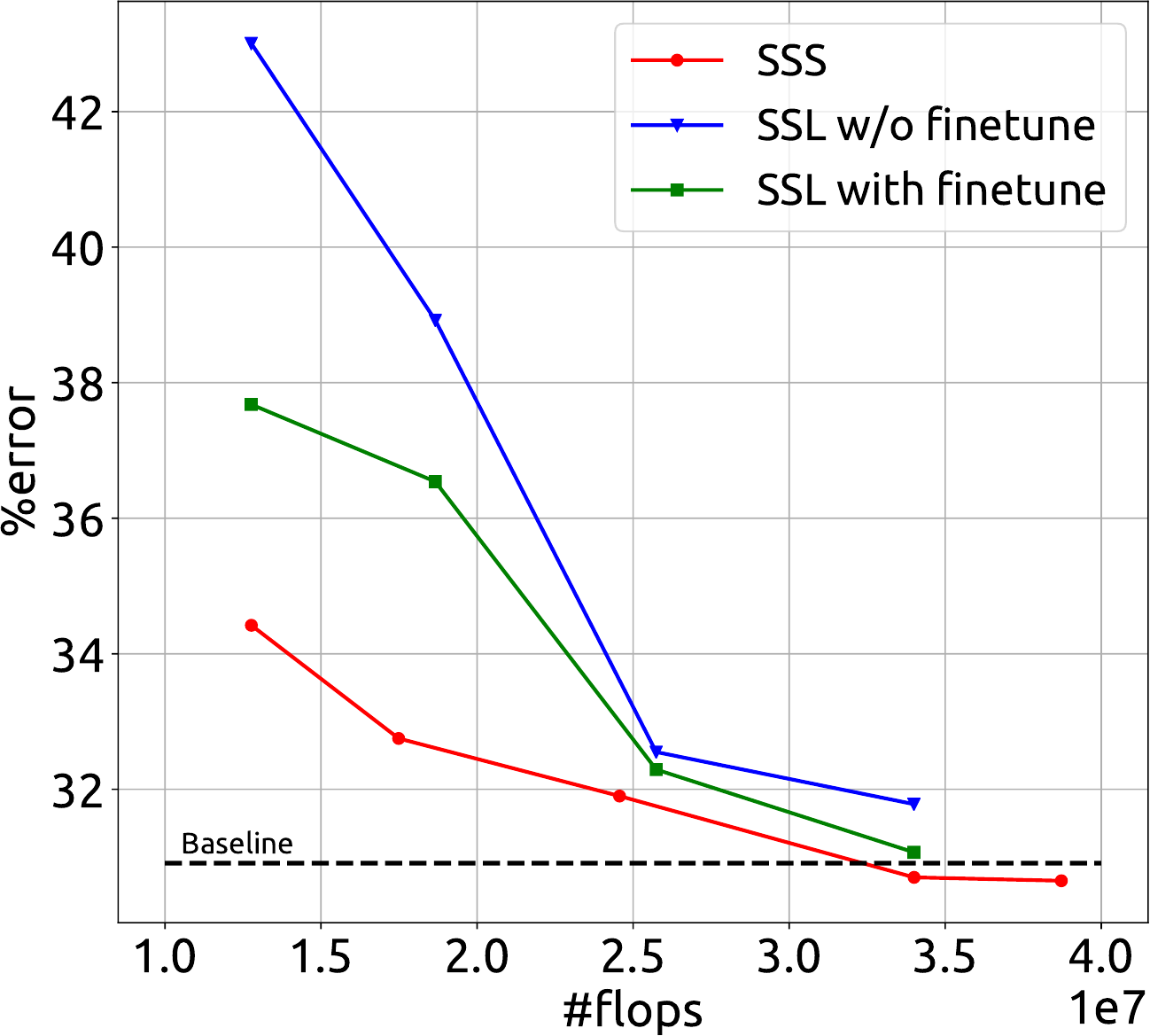}}
	\caption{Error vs. FLOPs for our SSS models and SSL models \label{fig:sslcifar}}
\end{figure*}

We compare our SSS with other pruning methods, including SSL \cite{wen2016learning}, filter pruning \cite{li2016pruning}, channel pruning \cite{he2017channel}, ThiNet \cite{luo2017thinet}, \cite{molchanovpruning} and \cite{ye2018rethinking}. 
We compare SSL with our method in CIFAR10 and CIFAR100. All the models are trained from scratch. As shown in Fig. \ref{fig:sslcifar}, our SSS achieves much better performances than SSL, even SSL with finetune.
Table \ref{table:resnet-comparison} shows the pruning results on the ImageNet LSVRC2012 dataset. To the best of our knowledge, only a few works reported ResNet pruning results with FLOPs. Comparing with filter pruning results, our ResNet-32 performs best with least FLOPs. As for channel pruning, with similar FLOPs\footnote{We calculate the FLOPs of He's models by provided network structures.}, our ResNet-32 yields 1.88\% lower top-1 error and 1.11\% lower top-5 error than pruned ResNet-50 provided by \cite{he2017channel}. As for \cite{ye2018rethinking}, our ResNet-41 achieves about 1\% lower top-1 error with less computation budge. We also show comparison in VGG16. All the method including channel pruning, ThiNet and our SSS achieve significant improvement than \cite{molchanovpruning}. Our VGG16 pruning result is competitive to other state-of-the-art.  

We further compare our pruned ResNeXt with DenseNet \cite{huang2016densely} in Table \ref{table:resnext-comparison}. With 14\% less FLOPs, Our ResNeXt-38 achieves 0.2\% lower top-5 error than DenseNet-121.
\begin{table}[t]
	\footnotesize
	\setlength{\tabcolsep}{4.5pt}
	\centering
	\caption{Comparison among several state-of-the-art pruning methods on the ResNet and VGG16 networks}	\label{table:resnet-comparison}
	%\vspace{8pt}
	\begin{tabular}{l|c|c|c}
		\toprule
		Model & Top-1 & Top-5 & \#FLOPs  \\ \midrule
		ResNet-34-pruned \cite{li2016pruning}& 27.44 & - & 3.080B  \\
		ResNet-50-pruned-A \cite{li2016pruning} (Our impl.)& 27.12 & 8.95 & 3.070B\\
		ResNet-50-pruned-B \cite{li2016pruning} (Our impl.)& 27.02 & 8.92 & 3.400B  \\ 
		ResNet-50-pruned (2$\times$) \cite{he2017channel} & 27.70 & 9.20 & 2.726B\\ 
		\textbf{ResNet-32 (Ours)} & \textbf{25.82} & \textbf{8.09} & \textbf{2.818B} \\ \midrule
		ResNet-101-pruned \cite{ye2018rethinking}& 25.44 & - & 3.690B \\
		\textbf{ResNet-41 (Ours)} & \textbf{24.56} & \textbf{7.39} & \textbf{3.473B} \\ \midrule
		VGG16-pruned \cite{molchanovpruning} & - & 15.5 & $\approx$8.0B \\
		VGG16-pruned (5$\times$) \cite{he2017channel} & 32.20 & 11.90 & 7.033B \\
		VGG16-pruned (ThiNet-Conv) \cite{luo2017thinet} & 30.20 & 10.47 & 9.580B \\
		VGG16-pruned (Ours) & 31.47 & 11.80 & 7.667B \\
		\bottomrule
	\end{tabular}
	
\end{table}
\begin{table}
	\footnotesize
	\setlength{\tabcolsep}{4.5pt}
	\centering
	\caption{Comparison between pruned ResNeXt-38 and DenseNet-121}
	\begin{tabular}{l|c|c|c}
		\toprule
		Model & Top-1 & Top-5 & \#FLOPs  \\ \midrule
		DenseNet-121 \cite{huang2016densely}& 25.02 & 7.71 & 2.834B\\
		DenseNet-121 \cite{huang2016densely} (Our impl.) &  25.58 & 7.89 & 2.834B  \\ 
		\textbf{ResNeXt-38} (\textbf{Ours}) &  \textbf{25.02} & \textbf{7.50} & \textbf{2.431B}  \\
		\bottomrule
	\end{tabular}
	%\vspace{-8pt}
	\label{table:resnext-comparison}
\end{table}
\subsection{Choice of different optimization methods}
We compare our APG with other different optimization methods for optimizing $\lambda$ in our ImageNet experiments, including SGD adopted in \cite{liu2017learning} and ISTA \cite{beck2009fast} used in \cite{ye2018rethinking}. We adopted ResNet-50 for block pruning and train it from scratch. The sparse penalty $\gamma$ is set to 0.005 for all optimization methods. 
\begin{table}[t]
	\footnotesize
	\setlength{\tabcolsep}{4.5pt}
	\centering
	\caption{Comparison between different optimization methods}\label{table:SGD-comparison}
	\begin{tabular}{l|c|c|c}
		\toprule
		Model & Top-1 & Top-5 & \#FLOPs  \\ \midrule
		ResNet-32-SGD & 26.46 & 8.41 & 2.726B\\
		ResNet-32-APG & 25.82 & 7.39 & 2.818B  \\ 
		\bottomrule
	\end{tabular}
	%\vspace{-8pt}
\end{table}

For SGD, since we can not get exact zero scale factor during training, a extra hyper-parameter -- hard threshold is need for the optimization of $\lambda$. In our experiment, we set it to $0.0001$. After training, we get a ResNet-32-SGD network. As show in Table \ref{table:SGD-comparison}, the performance of our ResNet-32-APG is better than ResNet-32-SGD.

For ISTA, we found the optimization of network could not converge. The reason is that the converge speed of ISTA for $\lambda$ optimization is too slow when training from scratch. Adopting ISTA can get reasonable results in CIFAR dataset. However, in ImageNet, it is hard to optimize the $\lambda$ to be sparse with small $\gamma$, and larger $\gamma$ will lead too many zeros in our experiments. \cite{ye2018rethinking} alleviated this problem by fine-tunning from a pretrained model. They also adopted $\lambda$-$\W$ rescaling trick to get an small $\lambda$ initialization.

Comparing to ISTA, Our APG can be seen as a modified version of an improved ISTA, namely FISTA \cite{beck2009fast}, which has been proved to be significantly better than ISTA in convergence. Thus the optimization of our method is effective and stable in both CIFAR and ImageNet experiments. The results described in Table \ref{table:resnet-comparison} also show the advantages of our APG method to ISTA. The performance of our trained ResNet-41 is better than ResNet-101-pruned provided by \cite{ye2018rethinking}.

%% file: sections/6-Conclusions.tex
\section{Conclusions}
\label{sec:conclusions}
In this paper, we have proposed a data-driven method, Sparse Structure Selection (SSS) to adaptively learn the structure of CNNs. In our framework, the training and pruning of CNNs is formulated as a joint  sparse regularized optimization problem. Through pushing the scaling factors which are introduced to scale the outputs of specific structures to zero, our method can remove the structures corresponding to  zero scaling factors. To solve this challenging optimization problem and adapt it into deep learning models, we modified the Accelerated Proximal Gradient method.
In our experiments, we demonstrate very promising pruning results on PeleeNet, VGG, ResNet and ResNeXt. We can adaptively adjust the depth and width of these CNNs based on budgets at hand and difficulties of each task. We believe these pruning results can further inspire the design of more compact CNNs. 

In future work, we plan to apply our method in more applications such as object detection. It is also interesting to investigate the use of more advanced sparse regularizers such as non-convex relaxations, and adjust the penalty based on the complexity of different structures adaptively.